\documentclass[letterpaper]{article} 
\usepackage{aaai2027}
\nocopyright
\usepackage[hyphens]{url}  
\usepackage{graphicx} 
\usepackage{natbib}  
\usepackage{caption} 
\usepackage{booktabs}
\usepackage{multirow}
\usepackage{amsmath}
\usepackage{amssymb}
\usepackage{algorithm}
\usepackage{algorithmic}
\title{When Correct Solutions Repeat: Rarity-Aware Credit Redistribution for GRPO}
\author{Zhe Cao, Miaowen Wen, Fangjiong Chen}
\affiliations{South China University of Technology}

\begin{document}

\captionsetup{justification=justified}

\maketitle


\begin{abstract}
Reinforcement learning with verifiable rewards (RLVR) commonly optimizes each correct completion as an independent learning signal. In GRPO, this completion-level uniformity creates structure-level skew: recurring correct solution forms accumulate positive coefficient mass in proportion to how often they are sampled, while rare forms receive limited credit. We formalize this behavior as multiplicity-induced structure-level credit concentration and introduce a partition-conditioned rule that redistributes positive advantages according to cluster rarity. Cue-GRPO instantiates this rule without auxiliary-model inference by using deterministic Strategy Cues to construct rollout-local partitions of verified-correct traces. Across Qwen2.5-Math-7B and Llama-3.1-8B-Instruct, Cue-GRPO improves AIME repeated-sampling performance, with the largest gains at high sampling budgets. Credit Redistribution (CR) under Judge Partitions (JP) further indicates that the proposed redistribution mechanism can operate with judge-derived partitions. Cue-GRPO adds only 6\% wall-clock training overhead over GRPO. These results support structure-level credit redistribution as a practical design axis for RLVR, with Strategy Cues providing a low-overhead implementation for competition mathematics. Code is available at \url{https://github.com/CzZ12/When-Correct-Solutions-Repeat-Rarity-Aware-Credit-Redistribution-for-GRPO}.
\end{abstract}


\section{Introduction}

Reinforcement learning with verifiable rewards (RLVR) has emerged as a powerful paradigm for improving the reasoning capabilities of large language models~\citep{deepseekr1,kimik15,openreasonerzero}. Group Relative Policy Optimization (GRPO; \citealt{grpo}) is widely used in this setting because it replaces a learned value critic with reward normalization over sampled completion groups. Under binary rewards, however, every verified-correct completion in a non-degenerate group receives the same positive base advantage. When several completions repeat the same underlying solution form, that form accumulates positive coefficient mass in proportion to its observed multiplicity, whereas a rare correct form receives only the credit of its few instances. GRPO is therefore uniform at the completion level but frequency-skewed after credit is aggregated by recurring structure. We refer to this behavior as multiplicity-induced structure-level credit concentration.

This allocation is consistent with expected-reward optimization, but it is poorly aligned with high-budget repeated sampling, where retaining alternative correct modes can improve the probability of solving difficult problems. Prior work addresses this setting through entropy regularization~\citep{cheng2026reasoning}, pass@$k$-oriented objectives~\citep{pkpo}, and set-level policy optimization~\citep{setpo}. Most closely, UARL~\citep{rewardingrare} uses an auxiliary LLM judge to group correct responses by high-level strategy and emphasizes rare groups. Its results show that rare solution groups can be useful during RLVR, but the resulting partition is entirely determined by the judge's outputs and requires substantial auxiliary-model inference throughout training. This motivates a more general view: partition construction and positive-credit redistribution are two distinct design components. The central question is how credit should be allocated once a set of verified-correct completions has been partitioned, and whether a useful partition can be obtained without full semantic strategy recovery.

We address this question by introducing Cue-GRPO, a low-overhead method that separates partition construction from credit allocation. At its core, a partition-conditioned Credit Redistribution (CR) rule rebalances positive advantages across groups of verified-correct completions while leaving negative advantages and the standard GRPO objective unchanged. Cue-GRPO constructs these groups using deterministic Strategy Cues extracted from observable mathematical operations and reasoning procedures, and then applies finite-group stabilization to the redistributed coefficients. In this design, CR determines how credit enters the policy update, while Strategy Cues provide the partition without online auxiliary-model inference.

Experiments with Qwen2.5-Math-7B and Llama-3.1-8B-Instruct show that Cue-GRPO improves AIME repeated-sampling performance, with the most reliable gains at high sampling budgets. Mechanism analysis confirms the intended redistribution: the empirical slope relating cluster size to normalized pre-floor positive coefficient mass decreases from 1.00 under GRPO to 0.72 under Cue-GRPO. To test whether CR depends on cue-derived partitions, we also evaluate Credit Redistribution under Judge Partitions (CR-JP), a control that uses the same 32B judge-based partitioning pipeline as our UARL baseline while applying the CR configuration to the resulting groups. On AIME, Cue-GRPO performs best while adding only $\sim$6\% training overhead over GRPO, compared with $\sim$62\% for the judge-based baseline. On HLE, CR-JP closely tracks UARL in overall AUC and attains a numerically higher pass@256, showing that CR can also operate with judge-derived partitions.

In summary, our main contributions are threefold:
\begin{itemize}
\item We identify and formalize multiplicity-induced structure-level credit concentration in GRPO: equal per-completion advantages cause recurring correct structures to accumulate positive coefficient mass in proportion to their observed frequency.
\item We develop a partition-conditioned, rarity-aware CR rule that reallocates positive credit across groups of verified-correct completions while leaving negative advantages unchanged. The CR-JP control further indicates that CR can operate with judge-derived partitions.
\item We instantiate the rule as Cue-GRPO, using deterministic Strategy Cues to construct rollout-local partitions without auxiliary-model inference, and demonstrate improved high-budget AIME performance across two model families with $\sim$6\% training overhead.
\end{itemize}


\section{Methodology}

\subsection{Structure-Level Credit Concentration in GRPO}

Given a prompt $x$, let $\pi_\theta$ denote the current policy parameterized by $\theta$. GRPO independently samples a rollout group of $K$ completions, $o_i\sim\pi_\theta(\cdot\mid x)$ for $i=1,\ldots,K$, where each completion is a full reasoning trace ending with a final answer. A rule-based verifier compares the extracted answer with the ground truth $y$ and assigns a binary reward $r_i\in\{0,1\}$. Let $\mu$ and $\sigma$ denote the mean and standard deviation of the group rewards. The group-relative advantage is $A_i=(r_i-\mu)/\sigma$. In our implementation, when $\sigma=0$, all normalized advantages are set to zero.

With binary rewards and nonzero reward variance, every verified-correct completion receives the same positive base advantage,
\begin{equation}
A^+=\frac{1-\mu}{\sigma}.
\end{equation}
This completion-level uniformity induces frequency-dependent allocation at the structure level. Let $S$ be an idealized set of verified-correct completions that instantiate the same recurring solution form, and let $n_S=|S|$. The aggregate positive advantage coefficient assigned to $S$ is
\begin{equation}
M_{\mathrm{GRPO}}(S)
=\sum_{i\in S}A_i
=n_SA^+.
\label{eq:linear_credit}
\end{equation}
Equation~(2) makes the source of concentration explicit: a recurring correct form receives more total positive credit simply because it appears more often in the rollout group. We address this effect by redistributing positive credit over a partition of the verified-correct completions, while leaving incorrect-completion advantages unchanged. We first define the redistribution rule for an arbitrary partition and then describe how Cue-GRPO constructs that partition.

\subsection{Partition-Conditioned Credit Redistribution}

Let $\mathcal{P}=\{i:r_i=1\}$ denote the verified-correct indices and $N=|\mathcal{P}|$. Consider any partition $\mathcal{C}=\{C_1,\ldots,C_m\}$ of $\mathcal{P}$, and let $C_i$ denote the cluster containing completion~$i$, with size $|C_i|$. Given a rarity exponent $\alpha\in[0,1]$, we redistribute the GRPO advantages as
\begin{equation}
A_i^{\mathrm{CR}}=
\begin{cases}
\displaystyle
A_i\,
\frac{N\,|C_i|^{-\alpha}}
{\displaystyle\sum_{j\in\mathcal{P}}|C_j|^{-\alpha}},
& i\in\mathcal{P},\\[10pt]
A_i,
& i\notin\mathcal{P}.
\end{cases}
\label{eq:cr_core}
\end{equation}

Denote the positive-completion multiplier by $\omega_i$, such that $A_i^{\mathrm{CR}}=\omega_iA_i$ for $i\in\mathcal{P}$. Completion normalization ensures $\frac{1}{N}\sum_{i\in\mathcal{P}}\omega_i=1$, so the core transformation changes the distribution of positive credit without changing its mean multiplier. Incorrect-completion advantages remain unchanged.

To characterize the resulting allocation, consider a cluster $C\in\mathcal{C}$ with size $n_C=|C|$. Under binary rewards, all verified-correct completions share the same positive base advantage $A^+$. The aggregate coefficient mass assigned to $C$ is
\begin{equation}
M_{\mathrm{CR}}(C)=\sum_{i\in C}A_i^{\mathrm{CR}}
=\frac{NA^+\,n_C^{\,1-\alpha}}
{\sum_{C'\in\mathcal{C}}n_{C'}^{\,1-\alpha}}.
\label{eq:cr_mass}
\end{equation}

Thus, $\alpha$ directly controls frequency compression. At the core redistribution stage, $\alpha=0$ recovers GRPO's allocation, whereas $\alpha=1$ assigns equal aggregate positive credit to all observed correct clusters. Intermediate values preserve a dependence on cluster frequency while reducing the linear concentration induced by GRPO. This rule is agnostic to how $\mathcal{C}$ is constructed; Cue-GRPO supplies a deterministic partition as described next.

\begin{figure*}[t]
\centering
\includegraphics[width=\textwidth]{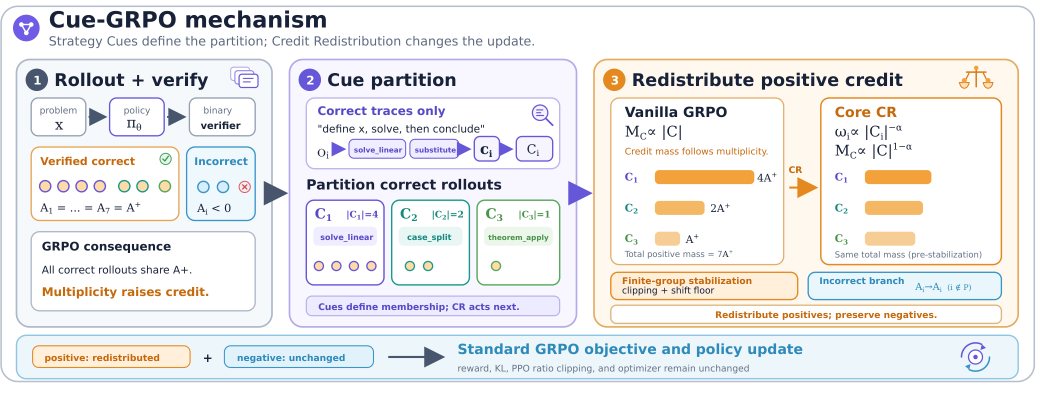}
\caption{Illustration of Cue-GRPO's credit-redistribution mechanism.
A binary verifier first separates correct and incorrect rollouts. Strategy cues partition verified-correct traces into cue-defined clusters, and each correct rollout is weighted by $|C_i|^{-\alpha}$. This changes cluster-level positive credit from $M_C \propto |C|$ in vanilla GRPO to $M_C \propto |C|^{1-\alpha}$, reducing multiplicity-driven concentration while preserving the total positive coefficient mass before finite-group stabilization. Incorrect rollouts retain their original negative advantages, and the resulting advantages enter the standard GRPO objective. The $4{:}2{:}1$ partition is illustrative.}
\label{fig:mechanism}
\end{figure*}

\begin{table}[t]
\centering
\setlength{\tabcolsep}{14pt}
\begin{tabular}{@{}ll@{}}
\toprule
Cue & Example triggers \\
\midrule
\multicolumn{2}{c}{\textit{Symbolic (\LaTeX)}} \\
\midrule
modulo\_op      & \verb|\pmod|, \verb|\bmod|, \verb|\equiv| \\
combinatorics   & \verb|\binom|, \verb|\choose| \\
radicals        & \verb|\sqrt|, \verb|\sqrt[]| \\
summation       & \verb|\sum|, \verb|\prod| \\
log\_exp        & \verb|\log|, \verb|\ln|, \verb|\exp| \\
trigonometry    & \verb|\sin|, \verb|\cos|, \verb|\tan| \\
inequality      & \verb|\leq|, \verb|\geq|, \verb|\neq| \\
\midrule
\multicolumn{2}{c}{\textit{Natural-language}} \\
\midrule
define          & let $x$ be, assume, suppose \\
substitute      & substitute, plug in, replace \\
solve\_linear   & solve, isolate, rearrange \\
case\_split     & Case~1, consider two cases, otherwise \\
theorem\_apply  & theorem, Pythagorean, Fermat \\
compute         & compute, calculate, evaluate \\
simplify        & simplify, reduce, cancel, combine \\
\bottomrule
\end{tabular}
\caption{Representative Strategy Cues used by Cue-GRPO. This table is not exhaustive; the complete fixed catalog of 27 cues is in Appendix~A.}
\label{tab:cues}
\end{table}

\subsection{Deterministic Partition Construction with Strategy Cues}

To instantiate Eq.~(3), Cue-GRPO requires a computable partition of the verified-correct completions. We find that recurring solution structures can be approximated from observable mathematical operations and procedural markers already present in the reasoning traces. Cue-GRPO therefore constructs the partition using deterministic Strategy Cues. The resulting cue signatures serve as operational indicators of recurring structure within each rollout group, rather than canonical semantic labels of proof strategy.

Our fixed vocabulary $\mathcal{L}$ contains 27 cues, including case splitting, substitution, modular reasoning, theorem application, and equation solving. For each trace $o_i$, the deterministic extractor $\Phi_0$ produces a cue sequence $\Phi_0(o_i)\in(\mathcal{L}\cup\{\texttt{other}\})^*$ before group-level filtering. Table~\ref{tab:cues} lists representative cues; Appendix~A gives the complete catalog. The catalog combines symbolic \LaTeX\ triggers, such as \texttt{\textbackslash sqrt}, \texttt{\textbackslash binom}, and \texttt{\textbackslash pmod}, with natural-language triggers, such as ``substitute,'' ``let $x$ be,'' and ``Case~1.'' Symbolic matches take precedence over natural-language matches; if several natural-language cues match the same step, a fixed catalog priority resolves the tie. The catalog was defined before training, kept fixed across backbones and evaluations, and not selected using any specific evaluation benchmark.

\subsection{Operational Cue Extraction and Clustering}

For each completion, the extraction pipeline: (1)~segments the trace into reasoning steps using paragraph boundaries and discourse markers; (2)~assigns the highest-priority matching cue to each step, using \texttt{other} when no cue matches; (3)~merges consecutive duplicate cues and removes the generic \texttt{verify} and \texttt{other} labels; and (4)~suppresses group-common cues. More precisely, let $\bar{s}_i$ denote the sequence after duplicate merging and generic-label removal, and for each $\lambda\in\mathcal{L}$ define
\begin{equation}
q_\lambda=\sum_{i\in\mathcal{P}}\mathbf{1}[\lambda\in\bar{s}_i],
\label{eq:q_lambda}
\end{equation}
where $\mathbf{1}[\cdot]$ is the indicator function. Given a suppression threshold $\rho\in[0,1]$, cue $\lambda$ is removed when $q_\lambda>\rho|\mathcal{P}|$. This makes the final cleaned skeleton $s_i=\Phi(o_i;\mathcal{P})$ explicitly dependent on the rollout group, while the underlying extractor and catalog remain deterministic.

We convert $s_i$ into a bag-of-cues count vector $\mathbf{c}_i\in\mathbb{Z}_{\ge 0}^{d}$, where $d=|\mathcal{L}|=27$, and each coordinate records the frequency of one catalog cue after group-level filtering. Suppressed cues contribute zero to their coordinates rather than reducing the dimension.

Let $\mathcal{P}_{\mathrm{nz}}=\{i\in\mathcal{P}:\|\mathbf{c}_i\|_2>0\}$ denote the verified-correct completions with nonempty cue vectors, and let $\mathcal{P}_0=\mathcal{P}\setminus\mathcal{P}_{\mathrm{nz}}$ contain those with no retained cues. For any $i,j\in\mathcal{P}_{\mathrm{nz}}$, we measure the similarity between their cue signatures using cosine similarity:
\begin{equation}
\operatorname{sim}_{\Phi}(i,j)=
\frac{\mathbf{c}_i^{\top}\mathbf{c}_j}
{\|\mathbf{c}_i\|_2\|\mathbf{c}_j\|_2}.
\label{eq:sim}
\end{equation}

Given a similarity threshold $\varepsilon\in[0,1]$, we construct an undirected graph $G_\Phi=(\mathcal{P}_{\mathrm{nz}},E_\Phi)$ with edge set $E_\Phi=\{\{i,j\}:i,j\in\mathcal{P}_{\mathrm{nz}},\,i<j,\,\operatorname{sim}_{\Phi}(i,j)\ge\varepsilon\}$. Let $\mathcal{K}_\Phi=\{K_{\Phi,1},\ldots,K_{\Phi,m_\Phi}\}$ denote the partition of $\mathcal{P}_{\mathrm{nz}}$ induced by the connected components of $G_\Phi$, where $m_\Phi$ is the number of connected components. All completions in $\mathcal{P}_0$ are assigned to a single \texttt{no-retained-cue} cluster, yielding the final Strategy-Cue partition
\begin{equation}
\mathcal{C}_\Phi=
\begin{cases}
\mathcal{K}_\Phi\cup\{\mathcal{P}_0\}, & \mathcal{P}_0\neq\varnothing,\\
\mathcal{K}_\Phi, & \mathcal{P}_0=\varnothing.
\end{cases}
\label{eq:partition}
\end{equation}

We instantiate the partition-conditioned rule in Eq.~(3) with $\mathcal{C}=\mathcal{C}_\Phi$, identifying each $C_i$ with the cue-derived cluster from $\mathcal{C}_\Phi$. This separation makes the role of Strategy Cues explicit: they determine which correct completions share a group, while Eq.~(3) determines how positive credit is redistributed across the resulting groups.

\subsection{Stabilized Cue-GRPO Advantages}

The normalized multipliers from Eq.~(3) define the core redistribution. To stabilize their use in finite rollout groups, multipliers associated with anomalous cue-specific singleton clusters are reset to one according to the criterion in Appendix~B. The remaining multipliers are clipped to $[\gamma_{\min},\gamma_{\max}]$; let $\bar{\omega}_i$ denote the resulting value. We then apply a minimum positive multiplier $\tau>0$:
\begin{equation}
\widehat A_i=
\begin{cases}
\displaystyle
A_i\left(
\bar{\omega}_i+
\bigl[\tau-\min_{j\in\mathcal{P}}\bar{\omega}_j\bigr]_+
\right), & i\in\mathcal{P},\\[8pt]
A_i, & i\notin\mathcal{P},
\end{cases}
\label{eq:final_advantage}
\end{equation}
where $[z]_+=\max(z,0)$.

\section{Experimental Setup}
\label{sec:exp_setup}

\subsection{Training Data and Optimization}

Our primary experiments fine-tune Qwen2.5-Math-7B~\citep{qwen2math} with LoRA~\citep{lora} on the first 1,000 Level~3--5 problems from MATH~\citep{hendrycks2021math}, accessed via the \texttt{qwedsacf/competition\_math} Hugging Face dataset. We retain only problem statements and final answers and discard reference solutions. Training spans one epoch over the 1,000-problem pool. For each prompt, the policy generates a rollout group of $K{=}64$ completions at temperature~1.0, with a maximum of 1,024 new tokens. Each group of 64 completions is consumed in optimization minibatches of eight completions with gradient accumulation~1, yielding 8,000 optimizer steps in total. Optimization uses paged AdamW 8-bit with learning rate $5{\times}10^{-7}$, a cosine schedule, policy-ratio clipping threshold 0.2~\citep{ppo}, and KL coefficient 0.001. LoRA uses rank 16, scaling 32, and zero dropout. Unless otherwise stated, training uses seed 42 on one NVIDIA H800 80\,GB GPU. We additionally fine-tune Llama-3.1-8B-Instruct~\citep{llama3} under the same data, rollout, and optimization protocol. Appendix~B reports the remaining implementation details.

\begin{table*}[ht]
\centering
\setlength{\tabcolsep}{5pt}
\begin{tabular}{@{}llccc ccc ccc c@{}}
\toprule
Family & Method & \multicolumn{3}{c}{AIME} & \multicolumn{3}{c}{HLE} & \multicolumn{3}{c}{MATH500} & GSM8K \\
& & \multicolumn{3}{c}{AUC@64 / 128 / 256} & \multicolumn{3}{c}{AUC@64 / 128 / 256} & \multicolumn{3}{c}{AUC@32 / 64 / 128} & AUC@32 \\
\midrule
\multirow{5}{*}{\shortstack{Qwen2.5-\\Math-7B}}
& Base    & 27.58 & 31.72 & 36.20 & 15.68 & 21.44 & 27.47 & 75.69 & 79.83 & 82.97 & 90.77 \\
& GRPO    & 30.62 & 34.99 & \underline{39.89} & 16.49 & 22.27 & 28.36 & 76.57 & 79.96 & 82.60 & 91.68 \\
& UARL & 30.48 & 34.32 & 38.25 & \textbf{18.38} & \textbf{24.41} & \textbf{30.50} & 76.92 & 80.59 & 83.70 & \underline{91.88} \\
& CR-JP & \underline{31.09} & \underline{35.10} & 39.84 & \underline{17.51} & \underline{23.71} & \underline{30.38} & \textbf{77.28} & \textbf{80.84} & \underline{83.72} & 91.81 \\
& Cue-GRPO & \textbf{32.06} & \textbf{37.00} & \textbf{42.75} & 17.13 & 23.06 & 28.98 & \underline{77.09} & \underline{80.75} & \textbf{83.75} & \textbf{92.66} \\
\bottomrule
\end{tabular}
\caption{AUC@K results for Qwen2.5-Math-7B. CR-JP uses the same 32B judge-based partitioning pipeline as UARL but replaces the weight computation with the CR rule. Bold indicates best within each column; underline indicates second-best.}
\label{tab:main_qwen}
\end{table*}

\begin{figure*}[ht]
\centering
\begin{minipage}[t]{0.32\textwidth}
\centering
\includegraphics[width=\textwidth]{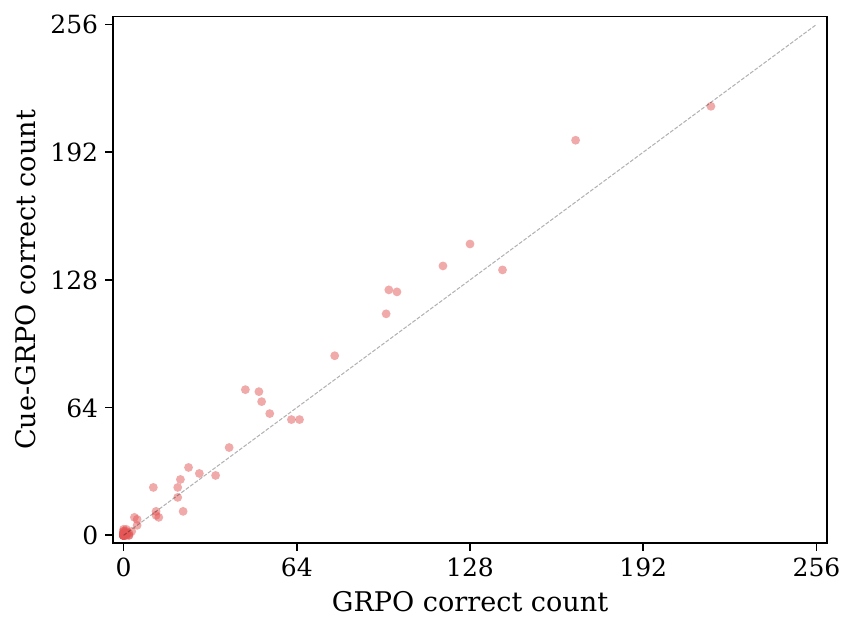}
\parbox{\textwidth}{\centering\footnotesize(a) Per-problem correct counts}
\end{minipage}\hfill
\begin{minipage}[t]{0.32\textwidth}
\centering
\includegraphics[width=\textwidth]{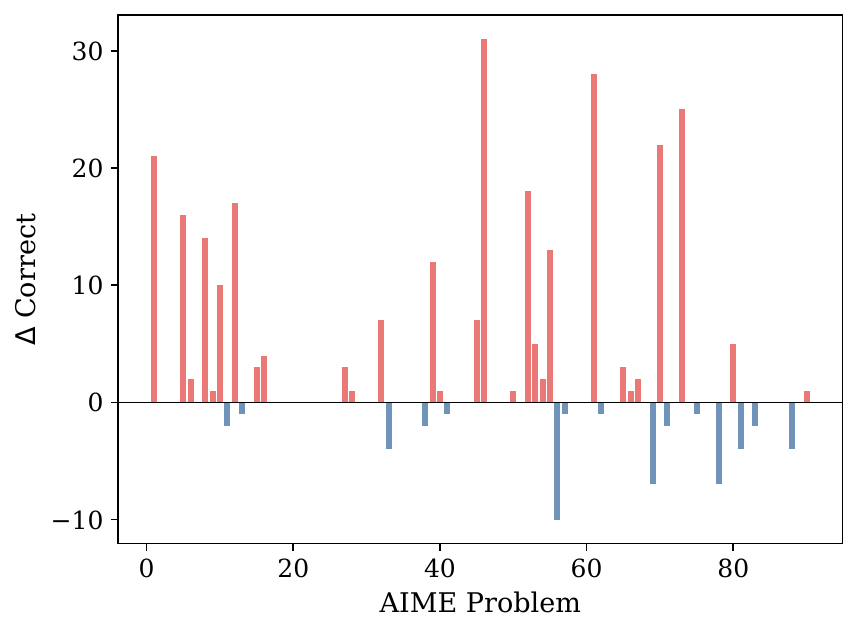}
\parbox{\textwidth}{\centering\footnotesize(b) Per-problem gain over GRPO}
\end{minipage}\hfill
\begin{minipage}[t]{0.32\textwidth}
\centering
\includegraphics[width=\textwidth]{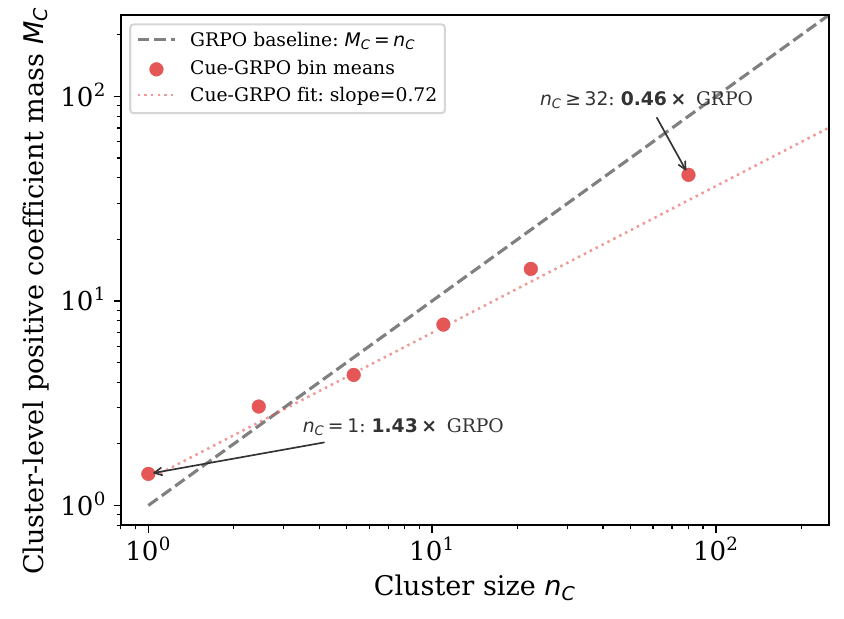}
\parbox{\textwidth}{\centering\footnotesize(c) Credit mass vs.\ cluster size}
\end{minipage}
\caption{Paired AIME correct-count gains and credit redistribution at $K{=}256$. (a) Cue-GRPO vs.\ GRPO per-problem correct counts; the diagonal indicates equal performance. (b) Per-problem paired gain $\Delta_p=c_p^{\mathrm{Cue}}-c_p^{\mathrm{GRPO}}$; red and blue bars indicate positive and negative gains. (c) Pre-floor diagnostic of the rarity-weighting transformation. In this panel, $M_C$ denotes the normalized pre-floor positive coefficient mass per unit advantage, $M_C=n_C\bar{\omega}_C$, where $\bar{\omega}_C$ is the cluster-level multiplier after artifact handling and clipping but before the additive shift floor.}
\label{fig:coverage}
\end{figure*}

\begin{table}[ht]
\centering
\small
\setlength{\tabcolsep}{3pt}
\begin{tabular}{@{}llccc c c@{}}
\toprule
Family & Method & \multicolumn{2}{c}{AIME} & MATH500 & GSM8K \\
& & \multicolumn{2}{c}{AUC@128 / 256} & AUC@64 & AUC@32 \\
\midrule
\multirow{4}{*}{\shortstack{Llama-3.1-\\8B-Instruct}}
& Base    & 17.13 & 22.99 & 69.91 & 91.41 \\
& GRPO    & \underline{18.85} & \underline{23.92} & \textbf{72.61} & \underline{92.95} \\
& UARL    & 17.94 & 23.11 & 71.32 & \textbf{93.00} \\
& Cue-GRPO & \textbf{19.77} & \textbf{25.71} & \underline{72.53} & 91.96 \\
\bottomrule
\end{tabular}
\caption{AUC@K results for Llama-3.1-8B-Instruct. Bold indicates best within each column; underline indicates second-best.}
\label{tab:main_llama}
\end{table}

\subsection{Compared Methods}

For each backbone, we compare five variants (Base, GRPO, UARL, CR-JP, and Cue-GRPO), with CR-JP evaluated on Qwen2.5-Math-7B only. Base is the initial checkpoint without RL fine-tuning. GRPO applies vanilla group-relative policy optimization with binary rewards. UARL is our compute-budgeted reimplementation of~\citet{rewardingrare}, adapted to our shared $K{=}64$ training recipe with a 4-bit Qwen2.5-32B-Instruct judge~\citep{qwen2} and $\alpha{=}0.5$.  CR-JP uses the same 32B judge-based partitioning pipeline as UARL but replaces UARL's weight computation with the CR rule ($\alpha{=}0.8$, clipping, shift floor); it is evaluated on Qwen2.5-Math-7B only. Cue-GRPO uses one fixed configuration across backbones: $\alpha{=}0.8$, $\varepsilon{=}0.50$, $\tau{=}1.05$, $\rho{=}0.75$, $\gamma_{\min}{=}0.3$, $\gamma_{\max}{=}3.0$.
All RL variants use the same 8,000-step training recipe. On the primary Qwen run, measured on the same H800, GRPO requires 4.7\,s per step, Cue-GRPO 5.0\,s, and the 32B-judge baseline 7.6\,s. These correspond to approximately 6\% and 62\% overhead over GRPO, respectively. Appendix~B gives the judge configuration, truncation, quantization, and weighting details.

\subsection{Benchmarks and Verification}

We evaluate repeated-sampling performance on AIME, our primary benchmark, which contains all 90 problems from AIME~I and II in 2022--2024. Normalized exact problem-text matching found no overlap between the 1,000-problem training pool and either AIME or MATH500. For Qwen2.5-Math-7B we also evaluate on MATH500 \citep{lightman2024lets} and GSM8K (fixed 200-problem test subset) as retention benchmarks~\citep{cobbe2021gsm8k}, and on HLE (fixed 200-problem subset of the mathematics split) as an out-of-distribution benchmark~\citep{phan2025hle}. Llama-3.1-8B-Instruct is evaluated on AIME, MATH500, and GSM8K.

Within each backbone, all methods use identical prompt templates, answer extraction, and verification. We extract the final boxed answer when present and otherwise use the last nonempty line, normalize both predictions and references, and apply conservative string-match verification without symbolic equivalence or numerical tolerance. Unparseable answers receive reward zero. All reported pass@$k$ values use the same pipeline across methods. Appendix~C gives the evaluation benchmarks, protocol, and answer extraction details.

\subsection{Repeated-Sampling Evaluation}

We generate $K_{\max}{=}256$ independent completions per problem for AIME and HLE, $K_{\max}{=}128$ for MATH500, and $K_{\max}{=}32$ for GSM8K. Evaluation uses vLLM~\citep{vllm} with temperature~1.0, top-$p{=}1.0$, and up to 2,048 new tokens per completion. Let $c_p$ denote the number of verified-correct completions among the $K_{\max}$ independent samples for problem $p$. We report $\mathrm{pass@}k$ using the standard unbiased estimator~\citep{chen2021codex}:
$\mathrm{pass@}k = \frac{1}{N}\sum_{p=1}^{N}\bigl[1 - \binom{K_{\max}-c_p}{k}/\binom{K_{\max}}{k}\bigr]$,
where $\binom{a}{b}=0$ for $a<b$; at $k=K_{\max}$, the estimator reduces to the fraction of problems with at least one verified-correct completion.

To summarize performance across sampling budgets, we compute the normalized trapezoidal area under the $\mathrm{pass@}k$ curve on the linear $k$-axis. For a reporting cap $K\leq K_{\max}$ and the ordered budgets $\{1,4,8,16,32,64,128,256\}\cap[1,K]$, we define
\begin{equation}
\mathrm{AUC@}K =
\frac{1}{K-1}
\sum_{j=1}^{m-1}
\frac{k_{j+1}-k_j}{2}
\bigl(p_j+p_{j+1}\bigr),
\label{eq:auc}
\end{equation}
where $p_j=\mathrm{pass@}k_j$. All reported methods use the same sampled-completion budgets and verification pipeline.


\section{Experiments}

\subsection{Main Results}

Tables~\ref{tab:main_qwen} and~\ref{tab:main_llama}, together with Fig.~\ref{fig:coverage}, evaluate three questions: whether structure-level CR improves repeated-sampling performance, whether the effect is tied to cue-derived partitions, and whether the realized coefficient allocation changes in the intended direction. The clearest gains appear on AIME, where single-sample accuracy is low and repeated correct forms are exposed at larger sampling budgets. On near-saturated benchmarks such as MATH500 and GSM8K, differences are smaller, consistent with the method targeting positive-credit allocation rather than uniformly increasing single-sample accuracy.

\noindent\textbf{Qwen2.5-Math-7B.}
Cue-GRPO improves over vanilla GRPO at all reported AIME budgets: AUC@64 increases from 30.62 to 32.06, AUC@128 from 34.99 to 37.00, and AUC@256 from 39.89 to 42.75. Cue-GRPO also outperforms the 32B-judge UARL baseline on AIME at all three budgets, with the largest margin at AUC@256 (42.75 vs.\ 38.25). The full pass@$k$ table shows that Cue-GRPO improves over GRPO at every evaluated AIME budget, with the gain increasing from $+1.0$ at $k{=}1$ to $+3.0$ at $k{=}128$ and $+4.4$ at $k{=}256$, confirming that the high-budget improvement is not obtained by sacrificing low-budget performance. At $K{=}256$, Cue-GRPO also increases problem coverage: it obtains at least one correct completion on 8 AIME problems where GRPO obtains none, while GRPO uniquely solves 4 problems where Cue-GRPO obtains none. Under our compute-budgeted training recipe, this result shows that a deterministic cue partition can provide an effective low-overhead implementation of Credit Redistribution, outperforming the 32B-judge rarity baseline on AIME without auxiliary-model inference.

\medskip
\noindent\textbf{Credit Redistribution under Judge Partitions.}
We further combine the proposed Credit Redistribution configuration with the same 32B judge-based partitioning pipeline used in our UARL reimplementation; we denote this control as CR-JP. On AIME, CR-JP reaches an AUC@256 of 39.84 and pass@256 of 47.8, compared with 38.25 and 44.4 for UARL, while Cue-GRPO further reaches 42.75 and 52.2. On HLE, CR-JP closely tracks UARL in AUC@256 (30.38 vs.\ 30.50) and attains a numerically higher pass@256 (40.5 vs.\ 39.5). These results provide complementary evidence that Credit Redistribution operates with judge-derived partitions, while the cue-based implementation provides the strongest AIME performance without online auxiliary-model inference. Full pass@$k$ results are reported in Appendix~D.
\medskip

\noindent\textbf{Llama-3.1-8B-Instruct.}
A similar AIME pattern holds: Cue-GRPO improves over GRPO from 18.85 to 19.77 at AUC@128 and from 23.92 to 25.71 at AUC@256, and is the best method on AIME among the compared variants. The gain is larger at the higher sampling budget, consistent with the hypothesis that structure-frequency skew becomes more consequential when more samples expose repeated correct solution forms. Since Llama is not initialized from a math-specialized checkpoint, this result suggests that the effect is not specific to a single model family. Under the same compute-budgeted comparison, Cue-GRPO remains above both GRPO and the 32B-judge rarity baseline on Llama AIME. Full pass@$k$ results are in Appendix~E.

\noindent\textbf{Retention and out-of-domain benchmarks.}
Taken together, the cross-benchmark results distinguish the general Credit Redistribution mechanism from its cue-based instantiation. Strategy Cues provide the strongest result on AIME, where the fixed catalog is well matched to competition-math reasoning, while pairing Credit Redistribution with judge-derived partitions attains the numerically highest pass@256 on HLE while remaining close to UARL in overall AUC. On MATH500 and GSM8K, the methods remain close as performance approaches saturation. Cue-GRPO therefore serves as an efficient, domain-matched implementation of a broader partition-conditioned redistribution principle.

\begin{table}[t]
\centering
\small
\setlength{\tabcolsep}{3pt}
\begin{tabular}{@{}lccc cc@{}}
\toprule
Method & Clustering & Rarity & Floor & \multicolumn{2}{c}{AIME} \\
& & & & \multicolumn{2}{c}{AUC@64/128} \\
\midrule
GRPO & --- & --- & --- & 29.18 & 32.64 \\
Uniform-Boost & --- & --- & $\checkmark$ & 29.45 & 33.13 \\
Random-Cluster & Random & $\checkmark$ & $\checkmark$ & 30.10 & 34.09 \\
Cue-GRPO ($\mu$-matched) & $\checkmark$ & $\checkmark$ & --- & 31.00 & 35.05 \\
Cue-GRPO ($\alpha{=}0.5$) & $\checkmark$ & $\checkmark$ & $\checkmark$ & \underline{31.22} & \underline{35.96} \\
\textbf{Cue-GRPO ($\alpha{=}0.8$)} & $\checkmark$ & $\checkmark$ & $\checkmark$ & \textbf{31.51} & \textbf{36.19} \\
Cue-GRPO ($\alpha{=}1.0$) & $\checkmark$ & $\checkmark$ & $\checkmark$ & 30.36 & 35.28 \\
\bottomrule
\end{tabular}
\caption{Ablation study on AIME ($K{=}128$, 90 problems). Models are independently trained.}
\label{tab:ablation}
\end{table}

\subsection{Coverage and Credit Redistribution Analysis}

Fig.~\ref{fig:coverage} decomposes the AIME result at the per-problem level for $K{=}256$. Fig.~\ref{fig:coverage}(a) compares the number of correct completions produced by Cue-GRPO and GRPO on each AIME problem. Most points lie on or near the diagonal, but the off-diagonal differences are skewed toward Cue-GRPO, indicating that the aggregate AIME gain is distributed across multiple problems rather than driven by a single outlier.

Fig.~\ref{fig:coverage}(b) makes this paired comparison explicit. For each problem $p$, we compute the correct-count difference $\Delta_p = c^{\mathrm{Cue}}_p - c^{\mathrm{GRPO}}_p$, where $c_p$ is the number of correct completions among 256 samples. Cue-GRPO has $\Delta_p>0$ on 29 problems, GRPO has $\Delta_p<0$ on 15 problems, and 46 problems are tied. Excluding ties, a two-sided exact sign test gives $p=0.049$, and the mean paired difference is $+2.52$ correct completions per problem (95\% bootstrap CI: $[1.14, 4.09]$). Of the 90 AIME problems at $K{=}256$, 39 are solved by both methods, 8 only by Cue-GRPO (12 correct completions total), 4 only by GRPO (6 correct completions), and 39 by neither. This asymmetry indicates that the high-budget gain reflects a modest expansion of the problem set reached by repeated sampling, rather than merely amplifying already-solved problems. Test definitions and resampling details are provided in Appendix~F.

Fig.~\ref{fig:coverage}(c) examines the rarity signal before the additive shift floor. Under GRPO, cluster-level positive coefficient mass grows linearly with cluster size (slope $1.00$). Cue-GRPO reduces the fitted log-log slope to $0.72$: singleton clusters receive $1.43\times$ the corresponding GRPO mass, whereas clusters in the largest displayed size range receive $0.46\times$. The empirical slope deviates from the theoretical pre-floor exponent $1-\alpha=0.2$ because clipping and singleton reset modify the realized weight distribution in finite rollout groups; the fitted slope therefore confirms that the implemented transformation reduces multiplicity-driven coefficient concentration before the additive floor is applied. The subsequent ablations test whether this change in credit allocation translates into repeated-sampling gains.

\subsection{Credit Redistribution Ablations}

Table~4 reports dedicated training runs evaluated at $K{=}128$; comparisons are therefore self-contained within this table and should not be made directly with Table~2. Full pass@$k$ curves are provided in Appendix~G.

Uniform-Boost multiplies every verified-correct advantage by $1.05$ and improves AUC@128 only modestly, from 32.64 to 33.13. Applying the minimum floor-level amplification uniformly is therefore insufficient to reproduce the full gain.

Random-Cluster improves AUC@128 from 32.64 to 34.09, indicating that partition-based CR can be beneficial even when membership is randomized. Cue-GRPO further raises AUC@128 to 36.19, a 2.10-point improvement over Random-Cluster. This additional gain is consistent with the benefit of organizing redistribution using recurring procedural cues.

Cue-GRPO ($\mu$-matched) removes the additive shift floor and re-normalizes the post-filter positive weights to unit mean. It therefore matches GRPO's mean positive-weight scale while retaining non-uniform redistribution. The variant reaches 35.05, preserving 68\% of the full improvement over GRPO and indicating that an increase in average positive-credit scale alone cannot account for the result. Restoring the full post-floor scaling raises AUC@128 from 35.05 to 36.19.

Across the tested rarity exponents, $\alpha{=}0.8$ performs best, indicating that moderate frequency compression provides the strongest balance in this setting. Overall, the ablations support non-uniform, partition-conditioned CR as an important contributor to the Cue-GRPO improvement, with cue-organized partitions providing the strongest result.

\subsection{Multi-Seed Robustness}

To assess training stability, we trained Cue-GRPO and GRPO with two additional seeds (7, 123). Table~\ref{tab:multi_seed} reports the AUC summary; full pass@$k$ tables are in Appendix~H. Across the three seeds, Cue-GRPO consistently improves AUC@256, with gains of $+1.90$, $+2.86$, and $+0.63$ (mean $+1.80$). At smaller budgets, it leads for two of the three seeds, yielding mean gains of $+0.53$ at AUC@64 and $+0.99$ at AUC@128. This pattern indicates that the benefit of Credit Redistribution is most reliable at high sampling budgets.

\begin{table}[t]
\centering
\setlength{\tabcolsep}{7pt}
\begin{tabular}{@{}lccc ccc@{}}
\toprule
Seed & \multicolumn{3}{c}{Cue-GRPO} & \multicolumn{3}{c}{GRPO} \\
& \multicolumn{3}{c}{AUC@64 / 128 / 256} & \multicolumn{3}{c}{AUC@64 / 128 / 256} \\
\midrule
7   & 30.90 & 34.98 & 39.48 & 30.31 & 33.90 & 37.58 \\
42  & 32.06 & 37.00 & 42.75 & 30.62 & 34.99 & 39.89 \\
123 & 29.92 & 33.92 & 38.66 & 30.37 & 34.04 & 38.03 \\
\bottomrule
\end{tabular}
\caption{Multi-seed AIME results (Qwen2.5-Math-7B, $K{=}256$).}
\label{tab:multi_seed}
\end{table}


\section{Related Work}

\noindent\textit{RLVR under repeated sampling and diversity-preserving optimization.}
Reinforcement learning with verifiable rewards (RLVR) has become a central paradigm for improving the reasoning capabilities of large language models~\citep{deepseekr1,kimik15,openreasonerzero}. Group Relative Policy Optimization (GRPO; \citealt{grpo}) is widely used in this setting because group-relative reward normalization removes the need for a learned value critic. However, improvements in single-sample accuracy do not always translate into stronger performance under repeated sampling. Prior analyses associate this gap with declining policy entropy and contraction of the model's output support during RLVR training~\citep{yue2025diversity,entropycollapse,cheng2026reasoning,strozzicrossover}, while CaSP~\citep{simko} directly studies candidate-support contraction and high-budget pass@$k$ degradation. Poly-EPO~\citep{polyepo} constructs set-level objectives that jointly reward accuracy and strategy diversity, providing positive learning signal even to novel-but-incorrect strategies. Cue-GRPO instead operates after verification, redistributing positive credit exclusively among verified-correct completions according to cue-defined structural frequency. Other GRPO variants primarily address optimization stability, training efficiency, or the treatment of easy and rare samples~\citep{drgrpo,simpleRLzoo,fgrpo,mcgrpo}.

A related line of work explicitly preserves exploration or optimizes repeated-sampling performance. Entropy-based methods encourage broader token-level support~\citep{cheng2026reasoning,huang2025}, pass@$k$-aware objectives directly optimize coverage across multiple samples~\citep{pkpo}, and set-level approaches coordinate groups of trajectories during policy optimization~\citep{setpo}. Multi-component systems such as DAPO and EDGE-GRPO also incorporate mechanisms intended to sustain exploration or diversify training signals~\citep{dapo,edgegrpo}. These approaches operate primarily on token distributions, sampling behavior, or trajectory sets. Our work instead studies the granularity of positive credit after verification: group-relative advantages treat correct completions uniformly even when several completions repeat the same solution structure. Cue-GRPO therefore complements exploration-oriented methods by redistributing credit within the set of already verified-correct completions, without directly rewarding novelty or optimizing pass@$k$.

\noindent\textit{Rarity-aware credit redistribution.}
DARLING~\citep{li2025darling} groups responses through learned semantic partitions and scales quality rewards using a diversity score that decreases with cluster frequency. UARL~\citep{rewardingrare} groups rollout sets by high-level strategy using an auxiliary LLM judge and scales GRPO advantages with inverse-power cluster-size weights. In contrast, we formulate credit allocation as a constrained, partition-conditioned reparameterization of positive GRPO coefficients. Given any explicit partition of the verified-correct subset, Credit Redistribution combines cluster-size-aware rarity shaping, positive-credit stabilization, and within-group mean restoration to transfer aggregate coefficient mass from overrepresented to underrepresented solution structures, while preserving the overall positive update scale and leaving negative advantages unchanged. Cue-GRPO instantiates this rule with deterministic Strategy Cues, while CR-JP applies it to judge-derived partitions.


\section{Conclusion}

We identify multiplicity-induced structure-level credit concentration in GRPO, revealing that equal per-completion advantages allocate disproportionate coefficient mass to dominant solution forms when they recur. We address this with partition-conditioned Credit Redistribution, which separates grouping from credit allocation. Cue-GRPO instantiates this formulation with deterministic Strategy Cues and delivers its clearest gains on AIME at high sampling budgets across two model families, with only $\sim$6\% overhead. The CR-JP control applies the same configuration to judge-derived partitions and exceeds the in-recipe UARL-32B baseline at pass@256 on both AIME and HLE. Together, these results identify partition-conditioned credit allocation as the broader principle: once recurring correct structures are exposed, their credit need not be tied to sample multiplicity. Strategy Cues provide a deterministic realization of this principle for competition mathematics; extending the same view to richer structural representations is a promising direction for broader reasoning domains.


\bibliography{refs}

\newpage
\onecolumn
\section{Appendix A. Complete Strategy Cue Catalog}

Table~\ref{tab:supp_full_cues} lists the complete set of 27 Strategy Cues used by Cue-GRPO. The main paper (Table~1) shows a representative subset of 14 cues.

\begin{table}[ht]
\centering
\small
\setlength{\tabcolsep}{3pt}
\begin{tabular}{@{}rll@{\hspace{8pt}}rll@{}}
\toprule
\multicolumn{3}{c}{Symbolic \LaTeX\ cues} & \multicolumn{3}{c@{}}{Natural-language cues} \\
\midrule
 1&modulo\_op      & \verb|\pmod|,\verb|\bmod|,\verb|\equiv|       & 13&define          & let $x$ be, assume, suppose, given \\
 2&combinatorics   & \verb|\binom|,\verb|\choose|,combination      & 14&equation\_setup & set up equation, we have the following \\
 3&discriminant    & \verb|\Delta|, $b^2-4ac$, discriminant       & 15&substitute      & substitute, plug in, replace \\
 4&radicals        & \verb|\sqrt|, \verb|\sqrt[]|                   & 16&solve\_linear   & solve, find, isolate, rearrange \\
 5&summation       & \verb|\sum|, \verb|\prod|                      & 17&solve\_quadratic& quadratic formula, complete square, factor \\
 6&calculus        & \verb|\int|,\verb|\lim|,\verb|\frac{d}|       & 18&solve\_cubic    & cubic equation, rational root theorem \\
 7&log\_exp        & \verb|\log|,\verb|\ln|,\verb|\exp|             & 19&case\_split     & Case~1, consider two cases, if \ldots otherwise \\
 8&trigonometry    & \verb|\sin|,\verb|\cos|,\verb|\tan|            & 20&theorem\_apply  & theorem, Pythagorean, Fermat, binomial \\
 9&inequality      & \verb|\leq|,\verb|\geq|,\verb|\neq|            & 21&modulo\_arith   & modulo, congruence, remainder, CRT \\
10&set\_theory    & \verb|\cup|,\verb|\cap|,\verb|\subset|         & 22&induction       & induction, base case, inductive hypothesis \\
11&matrix\_op     & \verb|\begin{matrix}|,\verb|\det|              & 23&combin\_count   & count, pigeonhole, stars and bars \\
12&gcd\_lcm       & \verb|\gcd|,\verb|\lcm|,greatest common        & 24&simplify        & simplify, reduce, cancel, combine \\
  &                &                                               & 25&compute         & compute, calculate, evaluate \\
  &                &                                               & 26&verify          & check, verify, test; \textit{filtered} \\
  &                &                                               & 27&conclude        & therefore, thus, \verb|\boxed| \\
\bottomrule
\end{tabular}
\caption{Complete catalog of 27 Strategy Cues used by Cue-GRPO. \texttt{verify} (\#26) and unmatched \texttt{other} are filtered before clustering. \texttt{conclude} (\#27) is retained but is often suppressed by within-group filtering when ubiquitous. Symbolic cues are matched before natural-language cues; ties among natural-language cues are resolved by the fixed catalog order (lower ID = higher priority). The catalog was defined before training and kept fixed across all models and benchmarks.}
\label{tab:supp_full_cues}
\end{table}

\begin{algorithm}[ht]
\caption{Cue-GRPO: deterministic partition construction and Credit Redistribution (per rollout group)}
\label{alg:supp_cue_grpo}

Algorithm~1 instantiates the partition-conditioned rule from the main paper. Steps~3--5 construct the Strategy-Cue partition, whereas Steps~6--13 compute and stabilize the redistributed positive advantages.

\textbf{Input:} completions $\{o_i\}_{i=1}^{K}$, rewards $\{r_i\}$, hyperparameters $\alpha,\varepsilon,\tau,\rho,\gamma_{\min},\gamma_{\max}$\\
\textbf{Output:} reweighted advantages $\{\widehat{A}_i\}$
\begin{algorithmic}[1]
\STATE Compute $\mu$ and $\sigma$ from $\{r_i\}$; \textbf{if} $\sigma = 0$ \textbf{then return} $\widehat{A}_i \leftarrow 0$ for all $i$.
\STATE Set $A_i \leftarrow (r_i-\mu)/\sigma$ and $\mathcal{P} \leftarrow \{i : r_i = 1\}$; \textbf{if} $\mathcal{P} = \emptyset$ \textbf{then return} $\widehat{A}_i \leftarrow A_i$ for all $i$.
\STATE Extract raw per-trace cue sequences $\Phi_0(o_i)$, merge consecutive duplicates, remove \texttt{verify} and \texttt{other}, and suppress cues occurring in ${>}\rho|\mathcal{P}|$ correct completions.
\STATE Convert the cleaned sequences for $i \in \mathcal{P}$ to bag-of-cues vectors $\mathbf{c}_i$; place all empty vectors in a single \texttt{no-retained-cue} cluster.
\STATE Build the cosine-threshold graph over nonempty vectors and compute its connected components; combine with the empty-vector cluster to form $\mathcal{C}$.
\FOR{each $C \in \mathcal{C}$}
    \STATE $n_C \leftarrow |C|$; $\bar{s}_C \leftarrow (1/n_C)^\alpha$.
\ENDFOR
\STATE Completion-normalize $\widetilde{s}_C \leftarrow \bar{s}_C \big/ \bigl(\frac{1}{n_c}\sum_{C'} n_{C'}\bar{s}_{C'}\bigr)$ where $n_c = \sum_{C'} n_{C'}$.
\STATE Clip to $[\gamma_{\min},\gamma_{\max}]$; reset artifact singletons to $1.0$.
\STATE Assign $w_i \leftarrow s_{C(i)}$ for $i \in \mathcal{P}$, $w_i \leftarrow 1$ otherwise.
\STATE $\delta \leftarrow \max(0,\;\tau - \min_{j\in\mathcal{P}} w_j)$; set $w'_i \leftarrow w_i + \delta$ for $i\in\mathcal{P}$, $w'_i \leftarrow 1$ otherwise.
\STATE \textbf{return} $\{\,A_i \cdot w'_i \mid i = 1,\dots,K\,\}$.
\end{algorithmic}
\end{algorithm}

\noindent\textbf{Extraction pipeline.} Completions are segmented into steps by splitting on paragraph boundaries, numbered lists (e.g., ``1. ''), and discourse markers (Step, First, Next, Then, Finally, Therefore, Thus, Hence, Now, We, Let, Consider, Suppose, Assume, Since, Because, etc.). Steps shorter than 20 characters are discarded. Each step is classified by the highest-priority matching cue (symbolic LaTeX patterns before natural-language text patterns). Consecutive duplicate cues are merged. The labels \texttt{verify} and \texttt{other} are removed as noise. Within each rollout group, any cue appearing in more than $\rho|\mathcal{P}|$ verified-correct completions ($\rho{=}0.75$) is suppressed to emphasize distinctions within the group. Empty skeletons after suppression form a single \texttt{no-retained-cue} cluster. Cosine similarity is computed on bag-of-cues count vectors; connected components of the threshold graph ($\varepsilon{=}0.50$) define the clusters. Singleton clusters with anomalously short or long skeletons are reset to weight 1.0 (thresholds in Appendix~B).

\noindent\textbf{Design rationale.} Each step has a specific motivation. Segmenting by paragraph boundaries and discourse markers separates distinct reasoning steps while keeping coherent operations together. Symbolic cues are matched before natural-language cues because LaTeX patterns are unambiguous. Filtering \texttt{verify} and \texttt{other} removes answer-format signals that do not reflect solution structure. Within-group cue suppression removes labels appearing in nearly all verified-correct completions, which provide no discriminatory signal. Connected-components clustering connects mutually similar completions without requiring all-pairs agreement, preventing fragmentation of near-identical traces while allowing chaining through intermediate forms. The bag-of-cues representation discards operation order for robustness to minor procedural rearrangements. These steps construct the operational partition used by the main paper's Credit Redistribution rule; the cue taxonomy itself does not determine the weighting function.

\section{Appendix B. Training and Implementation Details}

\noindent\textbf{Cue-GRPO.} We use Qwen2.5-Math-7B and Llama-3.1-8B-Instruct in bfloat16. LoRA: rank $r{=}16$, $\alpha_{\mathrm{LoRA}}{=}32$, dropout 0.0, task type \texttt{CAUSAL\_LM}, target modules q\_proj, k\_proj, v\_proj, o\_proj, gate\_proj, up\_proj, down\_proj (40.4M parameters for Qwen2.5-Math-7B). Optimizer: paged\_adamw\_8bit, lr $5{\times}10^{-7}$, weight decay 0.0, warmup ratio 0.0, max grad norm 1.0, cosine decay. Training spans one epoch over the fixed 1,000-problem pool (MATH Level~3--5). Prompt order is shuffled for the single training epoch; each prompt contributes one generated rollout group of $K{=}64$ completions (temperature~1.0, max 1,024 tokens each). Each group is consumed in optimization minibatches of eight completions with gradient accumulation~1, yielding 8,000 optimizer steps in total. KL penalty $\beta{=}0.001$ against the frozen initial checkpoint via GRPO clipped objective ($\epsilon{=}0.2$). Gradient checkpointing enabled (\texttt{use\_reentrant=False}). Seed 42 unless stated. Hardware: single NVIDIA H800 (80\,GB), vLLM 0.23.0, GPU memory 0.70, sleep mode, 4 dataloader workers. Prompt templates: Qwen uses \texttt{<|im\_start|>user} format, Llama uses \texttt{<|begin\_of\_text|>} format (activated via \texttt{USE\_LLAMA\_PROMPT=1}); both share the system instruction: ``You are a research mathematician. Solve the following problem step-by-step with rigorous reasoning. You MUST attempt to solve it no matter how difficult. Never refuse, never apologize, never mention the format. ALWAYS end with your final answer inside \texttt{\textbackslash boxed\{\}}.'' Checkpoints saved every 500 steps; at most 20 retained; evaluation at step 8,000. Cue-GRPO hyperparameters: $\alpha{=}0.8$, $\varepsilon{=}0.50$, $\tau{=}1.05$, $\rho{=}0.75$, $\gamma_{\min}{=}0.3$, $\gamma_{\max}{=}3.0$. Singleton artifact filtering: clusters of size 1 whose cleaned skeleton contains fewer than $\max(2,\,0.3{\times}\mathrm{median\_len})$ or more than $\max(6,\,3.0{\times}\mathrm{median\_len})$ cue labels are reset to $s_C{=}1.0$, where $\mathrm{median\_len}$ is the median skeleton length among verified-correct completions in the rollout group.

\noindent\textbf{GRPO.} Vanilla GRPO with binary rewards (correct ${=}1.0$, incorrect ${=}0.0$). All completions receive weight $w_i{=}1.0$, i.e., no rarity reweighting. Group-relative advantages are $A_i = (r_i - \mu)/\sigma$; if $\sigma = 0$ (all rewards identical), all advantages are set to zero. The same LoRA configuration, optimizer, $K{=}64$, temperature 1.0, max 1,024 tokens, KL $\beta{=}0.001$, gradient checkpointing, and random seed apply. For the GRPO baseline, cue extraction, clustering, and rarity reweighting are disabled, causing the implementation to fall back to standard GRPO advantage computation without calling the skeleton extraction or clustering pipeline.

\noindent\textbf{UARL.} Our compute-budgeted reimplementation uses a 4-bit quantized Qwen2.5-32B-Instruct judge with $\alpha{=}0.5$. After answer verification, the judge receives all $K{=}64$ completions per prompt, each truncated to 200 head and 500 tail characters. The system prompt is: ``You are an expert at analyzing mathematical reasoning strategies. Group the following solutions by HIGH-LEVEL MATHEMATICAL STRATEGY. Ignore superficial wording, variable names, verbosity. `factorization vs quadratic formula' = different. Same approach + more detail = same group. Return ONLY JSON: \texttt{\{"groups": [\{"strategy": "name", "members": [1,3]\}, \ldots]\}}.'' The judge output is parsed as JSON; if parsing fails, all completions are assigned to a single default group. Rarity frequencies are then computed from the judge-assigned labels of the verified-correct completions; incorrect completions retain weight~1. Correct-completion weights are computed as $w_C = (1/|C|)^\alpha$, normalized to have unit mean, without clipping or an additive floor. If at most one completion is correct, the judge is skipped and all weights default to~1. The judge is loaded via Transformers 4-bit quantization (BitsAndBytesConfig) alongside the 7B policy model and queried with max\_new\_tokens{=}512, do\_sample{=}False. The policy model's vLLM instance uses GPU memory utilization 0.45 to accommodate the additional judge. 
\medskip
\noindent\textbf{Credit Redistribution under Judge Partitions (CR-JP).}
This control uses the same 32B judge-based partitioning pipeline as UARL but replaces the UARL weighting configuration with the proposed Credit Redistribution configuration. Both variants compute inverse-frequency weights and normalize them over verified-correct completions; CR-JP uses $\alpha{=}0.8$, clipping to $[0.3,3.0]$, and the additive shift floor ($\tau{=}1.05$). Incorrect completions retain weight~1, and the cue-specific singleton safeguard is not applied because the judge pipeline does not produce cue skeletons.

\medskip
\noindent\textbf{Hardware and software.} All experiments run on a single NVIDIA H800 PCIe 80\,GB GPU with CUDA~13.0 and driver~580.82.07, under Ubuntu 22.04.5 LTS on an Intel Xeon Platinum 8458P with 1.0\,TiB system memory. Software versions: Python~3.12.3, PyTorch~2.11.0, Transformers~5.12.1, TRL~1.6.0, PEFT~0.19.1, vLLM~0.23.0, and BitsAndBytes~0.49.2.
\section{Appendix C. Evaluation Protocol and Verification}

\noindent\textbf{Evaluation benchmarks.} AIME: all 90 problems from AIME 2022--2024 (AIME~I and II for each year). MATH500: the 500-problem evaluation set introduced by~\citet{lightman2024lets}. GSM8K: the first 200 problems from the GSM8K test split~\citep{cobbe2021gsm8k}. HLE: the first 200 text-only problems from the math split of HLE~\citep{phan2025hle} (indices 0--199 of 976 total); the subset contains no problems requiring image inputs. Both fixed subset files are included in the submitted Code and Data Supplement.

\noindent\textbf{Evaluation protocol.} All evaluations use vLLM 0.23.0 with tensor parallelism~1, temperature~1.0, top-$p{=}1.0$, and up to 2,048 new tokens per completion. Completions are sampled i.i.d.\ for each problem without a fixed generation seed. Prompt templates are identical to training (see Appendix~B). For each benchmark, we generate $K_{\max}$ completions per problem (AIME and HLE: $K_{\max}{=}256$; MATH500: $K_{\max}{=}128$; GSM8K: $K_{\max}{=}32$). Let $c_p$ denote the number of verified-correct completions among the $K_{\max}$ independent samples for problem $p$. We report $\mathrm{pass@}k$ using the standard unbiased estimator~\citep{chen2021codex}:
\begin{equation}
\mathrm{pass@}k = \frac{1}{N}\sum_{p=1}^{N}\Bigl[1 - \binom{K_{\max}-c_p}{k}\Big/\binom{K_{\max}}{k}\Bigr],
\label{eq:passk_estimator}
\end{equation}
with the convention $\binom{a}{b}=0$ for $a<b$. At $k=K_{\max}$, this reduces to the observed fraction of problems with at least one correct completion. For a reporting cap $K\leq K_{\max}$, AUC@$K$ is computed via the normalized trapezoidal rule over the $\mathrm{pass@}k_j$ estimates (Eq.~9 in the main paper).

\noindent\textbf{Answer extraction and verification.} For all math benchmarks, we extract the final answer from the last \texttt{\textbackslash boxed\{\}} using balanced brace matching; if no boxed expression is found, we fall back to the last non-empty line. Extracted answers are normalized by stripping LaTeX wrappers, removing leading variable assignments and \% signs, normalizing whitespace, lowercasing, replacing ``and'' with commas, and deduplicating commas. Ground-truth answers are identically normalized. We adopt a conservative string-match verifier shared across all methods: extracted answers are compared via exact string equality after normalization, without symbolic equivalence or numeric tolerance. Because all methods use the same pipeline, relative rankings among methods are preserved. Unparseable answers receive reward~0.

\section{Appendix D. Full Pass@$k$ Results: Qwen2.5-Math-7B}

Table~\ref{tab:supp_qwen_all} reports complete pass@$k$ for Qwen2.5-Math-7B. AUC values for Base, GRPO, UARL, CR-JP, and Cue-GRPO are reported in Table~2 of the main paper.

\begin{table}[t]
\centering
\setlength{\tabcolsep}{3pt}
\begin{tabular}{@{}llcccccccc@{}}
\toprule
Dataset & Method & pass@1 & pass@4 & pass@8 & pass@16 & pass@32 & pass@64 & pass@128 & pass@256 \\
\midrule
\multirow{5}{*}{AIME}
& Base    & 6.3 & 15.2 & 20.5 & 25.3 & 29.6 & 33.6 & 38.0 & 43.3 \\
& GRPO    & 7.6 & 17.6 & 23.1 & 28.3 & 32.8 & 36.9 & 41.7 & 47.8 \\
& UARL    & 8.4 & 18.3 & 23.5 & 28.3 & 32.6 & 36.3 & 39.9 & 44.4 \\
& CR-JP & 8.8 & 19.1 & 24.4 & 29.1 & 33.1 & 36.8 & 41.3 & 47.8 \\
& Cue-GRPO & 8.6 & 18.9 & 24.3 & 29.4 & 34.1 & 39.0 & 44.7 & 52.2 \\
\midrule
\multirow{5}{*}{HLE}
& Base    & 1.1 & 4.0 & 7.0 & 11.4 & 17.2 & 23.8 & 30.4 & 36.5 \\
& GRPO    & 1.2 & 4.4 & 7.7 & 12.3 & 18.1 & 24.6 & 31.3 & 37.5 \\
& UARL    & 1.4 & 5.1 & 8.8 & 13.9 & 20.2 & 27.1 & 33.6 & 39.5 \\
& CR-JP & 1.4 & 4.7 & 8.2 & 13.1 & 19.2 & 26.1 & 33.5 & 40.5 \\
& Cue-GRPO & 1.3 & 4.6 & 8.0 & 12.7 & 18.8 & 25.6 & 32.2 & 37.5 \\
\midrule
\multirow{5}{*}{MATH500}
& Base    & 39.5 & 66.4 & 74.1 & 79.1 & 82.6 & 85.1 & 87.0 & -- \\
& GRPO    & 46.7 & 69.7 & 75.4 & 79.2 & 82.1 & 84.4 & 86.0 & -- \\
& UARL    & 47.2 & 69.8 & 75.5 & 79.5 & 82.8 & 85.5 & 88.0 & -- \\
& CR-JP & 44.7 & 70.1 & 76.2 & 80.1 & 83.1 & 85.5 & 87.6 & -- \\
& Cue-GRPO & 47.0 & 69.7 & 75.6 & 79.8 & 83.0 & 85.6 & 87.8 & -- \\
\midrule
\multirow{5}{*}{GSM8K}
& Base    & 45.1 & 80.7 & 90.5 & 95.4 & 97.0 & -- & -- & -- \\
& GRPO    & 54.2 & 84.0 & 91.0 & 95.1 & 97.5 & -- & -- & -- \\
& UARL    & 57.1 & 85.0 & 91.3 & 95.1 & 97.0 & -- & -- & -- \\
& CR-JP & 53.9 & 84.0 & 91.3 & 95.3 & 97.5 & -- & -- & -- \\
& Cue-GRPO & 54.9 & 86.1 & 92.8 & 96.0 & 97.5 & -- & -- & -- \\
\bottomrule
\end{tabular}
\caption{Full pass@$k$ (\%) for Qwen2.5-Math-7B. AUC values for all methods are reported in Table~2 of the main paper.}
\label{tab:supp_qwen_all}
\end{table}

\noindent\textbf{Interpretation.}
On AIME, Cue-GRPO improves over GRPO at every evaluated budget; the gap widens from $+1.0$ at pass@1 to $+4.4$ at pass@256 (AUC@256: $+2.86$), consistent with the intended effect of Credit Redistribution on high-budget coverage. CR-JP reaches 47.8 at pass@256 and 39.84 at AUC@256, compared with 44.4 and 38.25 for UARL. Cue-GRPO remains strongest at 52.2 and 42.75.

On HLE, CR-JP attains a numerically higher pass@256 than UARL (40.5 vs.\ 39.5; 81 vs.\ 79 of 200 problems), while the two methods remain closely matched in AUC@256 (30.38 vs.\ 30.50). Their curves are similar through $k{=}128$, with CR-JP moving ahead only at the final sampling budget. Together, these results suggest that the Credit Redistribution configuration can be paired with judge-derived partitions, while the relative advantage of each partition constructor varies across benchmarks.

On MATH500 and GSM8K, all methods are close, consistent with the smaller headroom for credit redistribution on benchmarks where performance approaches saturation.

\section{Appendix E. Full Pass@$k$ Results: Llama-3.1-8B-Instruct}

Table~\ref{tab:supp_llama_all} reports complete pass@$k$ for Llama-3.1-8B-Instruct, corresponding to Table~3 in the main paper.

\begin{table}[ht]
\centering
\setlength{\tabcolsep}{4pt}
\begin{tabular}{@{}llcccccccc@{}}
\toprule
Dataset & Method & pass@1 & pass@4 & pass@8 & pass@16 & pass@32 & pass@64 & pass@128 & pass@256 \\
\midrule
\multirow{4}{*}{AIME}
& Base    & 1.1 & 3.7 & 6.2 & 9.4 & 13.3 & 18.5 & 25.1 & 32.5 \\
& GRPO    & 1.9 & 5.9 & 8.9 & 12.2 & 15.8 & 20.2 & 25.7 & 32.2 \\
& UARL    & 1.9 & 5.8 & 8.8 & 11.9 & 15.2 & 19.1 & 24.3 & 32.2 \\
& Cue-GRPO & 1.8 & 5.7 & 8.7 & 12.2 & 16.2 & 21.2 & 27.6 & 35.6 \\
\midrule
\multirow{4}{*}{MATH500}
& Base    & 26.1 & 48.9 & 58.7 & 67.0 & 73.8 & 79.2 & -- & -- \\
& GRPO    & 32.1 & 54.0 & 62.8 & 70.1 & 76.1 & 80.8 & -- & -- \\
& UARL    & 32.0 & 53.5 & 62.2 & 69.3 & 74.8 & 78.6 & -- & -- \\
& Cue-GRPO & 31.4 & 53.6 & 62.7 & 70.2 & 76.1 & 80.6 & -- & -- \\
\midrule
\multirow{4}{*}{GSM8K}
& Base    & 43.4 & 80.6 & 90.9 & 96.1 & 98.5 & -- & -- & -- \\
& GRPO    & 56.8 & 86.7 & 92.8 & 96.0 & 98.0 & -- & -- & -- \\
& UARL    & 55.9 & 86.7 & 92.9 & 96.2 & 98.0 & -- & -- & -- \\
& Cue-GRPO & 55.1 & 85.5 & 91.7 & 94.9 & 97.5 & -- & -- & -- \\
\bottomrule
\end{tabular}
\caption{Full pass@$k$ (\%) for Llama-3.1-8B-Instruct. AUC values are reported in Table~3 of the main paper.}
\label{tab:supp_llama_all}
\end{table}

\noindent\textbf{Interpretation.} On Llama-3.1-8B-Instruct AIME, Cue-GRPO is slightly below GRPO at pass@1 ($1.8$ vs.\ $1.9$) but surpasses it from $k{=}32$ onward and reaches $+3.4$ at pass@256. This supports the claim that Cue-GRPO primarily improves high-budget repeated sampling rather than single-sample accuracy. On GSM8K, Cue-GRPO is below GRPO across budgets, indicating the method is not uniformly beneficial on saturated grade-school arithmetic tasks.

\section{Appendix F. Paired AIME Significance Tests}

For Fig.~2(b) in the main paper, let $c_p^{\mathrm{Cue}}$ and $c_p^{\mathrm{GRPO}}$ denote the number of correct completions among $K{=}256$ samples for AIME problem $p$, and define $\Delta_p=c_p^{\mathrm{Cue}}-c_p^{\mathrm{GRPO}}$. Across the 90 AIME problems, Cue-GRPO produces more correct completions on 29 problems, GRPO on 15, and 46 are tied. Excluding ties, a two-sided exact sign test gives $p=0.049$. The mean paired difference is $\bar{\Delta}=+2.52$, with a 95\% paired bootstrap CI of $[1.14,\,4.09]$ (10,000 paired resamples of the 90 problem-level pairs).

\section{Appendix G. Ablation Pass@$k$ Results}

All ablation models are independently trained from Qwen2.5-Math-7B (seed~42) with the same hyperparameters, evaluated on the same 90 AIME problems at $K{=}128$.

\noindent\textbf{Uniform-Boost.} Multiplies all verified-correct advantages by the shift floor constant ($\tau{=}1.05$) without clustering or rarity reweighting. Tests whether the minimum uniform boost associated with the floor is sufficient to reproduce the gain.

\noindent\textbf{Random-Cluster.} For each rollout group, let $m$ denote the number of cluster IDs returned by cue-based clustering. This control replaces cue-based membership by assigning each verified-correct completion an independently and uniformly sampled label from $\{1,\ldots,m\}$, and then applies the same rarity weighting, completion normalization, clipping, singleton safeguard, and shift floor as Cue-GRPO.

\noindent\textbf{Cue-GRPO ($\mu$-matched).} This variant removes the additive shift floor and re-normalizes the post-filter positive weights to unit mean, thereby matching GRPO's mean positive-weight scale while retaining non-uniform redistribution.

Table~\ref{tab:supp_ablation} reports full pass@$k$ curves corresponding to Table~4 in the main paper.

\begin{table}[ht]
\centering
\setlength{\tabcolsep}{9.5pt}
\begin{tabular}{@{}lccccccc@{}}
\toprule
Method & pass@1 & pass@4 & pass@8 & pass@16 & pass@32 & pass@64 & pass@128 \\
\midrule
GRPO & 7.7 & 17.4 & 22.8 & 27.5 & 31.2 & 34.3 & 37.8 \\
Uniform-Boost & 8.7 & 18.5 & 23.4 & 27.7 & 31.3 & 34.6 & 38.9 \\
Random-Cluster & 7.9 & 17.8 & 23.1 & 27.9 & 32.2 & 36.0 & 40.0 \\
Cue-GRPO ($\mu$-matched) & 8.4 & 18.4 & 23.9 & 28.9 & 33.1 & 37.0 & 41.1 \\
Cue-GRPO ($\alpha{=}0.5$) & 8.5 & 18.6 & 23.8 & 28.6 & 33.2 & 37.9 & 43.3 \\
Cue-GRPO ($\alpha{=}0.8$) & 8.1 & 18.3 & 23.8 & 28.9 & 33.6 & 38.3 & 43.3 \\
Cue-GRPO ($\alpha{=}1.0$) & 8.3 & 18.0 & 23.1 & 27.8 & 32.3 & 36.9 & 43.3 \\
\bottomrule
\end{tabular}
\caption{Full pass@$k$ (\%) for ablation experiments on Qwen2.5-Math-7B AIME ($K{=}128$). AUC values are reported in Table~4 of the main paper.}
\label{tab:supp_ablation}
\end{table}

\noindent\textbf{Interpretation.} Uniform-Boost changes the curve only modestly, showing that uniform $1.05\times$ scaling does not reproduce the full improvement. Random-Cluster improves mainly at larger $k$, indicating that randomized partition-based redistribution can already provide a useful high-budget training signal. Cue-GRPO produces the strongest curve, and its advantage over Random-Cluster grows from 0.2 points at pass@1 to 3.3 points at pass@128, consistent with the benefit of organizing credit according to recurring procedural cues. The mean-matched variant retains most of the gain while fixing the average positive-weight scale, and the $\alpha$ sweep favors the moderate setting $\alpha{=}0.8$.

\section{Appendix H. Multi-Seed Robustness}

Table~\ref{tab:supp_multi_seed} reports full pass@$k$ for Cue-GRPO and GRPO across three independent training seeds.

\begin{table}[ht]
\centering
\setlength{\tabcolsep}{6pt}
\begin{tabular}{@{}rllcccccccc@{}}
\toprule
Seed & Method & pass@1 & pass@4 & pass@8 & pass@16 & pass@32 & pass@64 & pass@128 & pass@256 \\
\midrule
\multirow{2}{*}{7}
& Cue-GRPO & 8.3 & 18.5 & 23.9 & 28.8 & 33.0 & 36.8 & 41.2 & 46.7 \\
& GRPO     & 8.4 & 18.5 & 23.7 & 28.4 & 32.4 & 35.7 & 39.1 & 43.3 \\
\midrule
\multirow{2}{*}{42}
& Cue-GRPO & 8.6 & 18.9 & 24.3 & 29.4 & 34.1 & 39.0 & 44.7 & 52.2 \\
& GRPO     & 7.6 & 17.6 & 23.1 & 28.3 & 32.8 & 36.9 & 41.7 & 47.8 \\
\midrule
\multirow{2}{*}{123}
& Cue-GRPO & 8.2 & 17.9 & 23.1 & 27.9 & 31.9 & 35.7 & 40.1 & 46.7 \\
& GRPO     & 8.7 & 18.9 & 24.1 & 28.5 & 32.3 & 35.8 & 39.5 & 44.4 \\
\bottomrule
\end{tabular}
\caption{Multi-seed AIME full pass@$k$ (\%) for Qwen2.5-Math-7B, $K{=}256$. Seed~42 is the main experiment.}
\label{tab:supp_multi_seed}
\end{table}

Across the three seeds, Cue-GRPO leads GRPO at pass@256 in all three seeds ($+3.4$ at seed~7, $+4.4$ at seed~42, $+2.3$ at seed~123). This variance is expected for RLVR fine-tuning and motivates reporting both paired per-problem statistics and multi-seed results.

\section{Appendix I. Training Dynamics and Runtime}

\noindent\textbf{Training dynamics.}
Zero-advantage groups increase from 0.0\% at training start to 9.6\% at training end as the model saturates on training problems. Over the training run, the operational cue partitions contain a mean of 2.25 clusters per activated rollout group; 91.8\% of groups have nonzero reward variance, 88.6\% contain multiple clusters, and 84.9\% receive non-uniform weights.

\noindent\textbf{Runtime comparison.}
Wall-clock training time is measured on a single NVIDIA H800 (80\,GB) under the same 8,000-step Qwen2.5-Math-7B recipe. GRPO averages 4.7\,s per step, Cue-GRPO 5.0\,s per step (approximately 6\% overhead), and UARL 7.6\,s per step (approximately 62\% overhead).

\section{Appendix J. Rarity Weighting Sensitivity}

Table~\ref{tab:supp_sensitivity} reports post-hoc sensitivity of the weight distribution.

\begin{table}[t]
\centering
\small
\setlength{\tabcolsep}{4pt}
\begin{tabular}{@{}lccccccc@{}}
\toprule
Config & $\varepsilon$ & $\alpha$ & $\rho$ & $\tau$ & Clip & $n_{\text{clust}}$ & $w_{\text{spread}}$ \\
\midrule
default                        & 0.50 & 0.80 & 0.75 & 1.05 & [0.3, 3.0] & 1.64 & 0.71 \\
\hline
$\varepsilon{=}0.30$           & 0.30 & 0.80 & 0.75 & 1.05 & [0.3, 3.0] & 1.50 & 0.71 \\
$\varepsilon{=}0.40$           & 0.40 & 0.80 & 0.75 & 1.05 & [0.3, 3.0] & 1.50 & 0.71 \\
$\varepsilon{=}0.60$           & 0.60 & 0.80 & 0.75 & 1.05 & [0.3, 3.0] & 2.21 & 1.04 \\
$\varepsilon{=}0.70$           & 0.70 & 0.80 & 0.75 & 1.05 & [0.3, 3.0] & 2.86 & 1.28 \\
\hline
$\alpha{=}0.3$                 & 0.50 & 0.30 & 0.75 & 1.05 & [0.3, 3.0] & 1.64 & 0.30 \\
$\alpha{=}0.5$                 & 0.50 & 0.50 & 0.75 & 1.05 & [0.3, 3.0] & 1.64 & 0.53 \\
$\alpha{=}1.0$                 & 0.50 & 1.00 & 0.75 & 1.05 & [0.3, 3.0] & 1.64 & 0.81 \\
\hline
$\rho{=}0.50$                  & 0.50 & 0.80 & 0.50 & 1.05 & [0.3, 3.0] & 2.50 & 0.95 \\
$\rho{=}0.90$                  & 0.50 & 0.80 & 0.90 & 1.05 & [0.3, 3.0] & 1.36 & 0.35 \\
\hline
$\tau{=}1.00$                  & 0.50 & 0.80 & 0.75 & 1.00 & [0.3, 3.0] & 1.64 & 0.71 \\
$\tau{=}1.10$                  & 0.50 & 0.80 & 0.75 & 1.10 & [0.3, 3.0] & 1.64 & 0.71 \\
$\tau{=}1.20$                  & 0.50 & 0.80 & 0.75 & 1.20 & [0.3, 3.0] & 1.64 & 0.71 \\
\hline
clip $[0.1,5.0]$               & 0.50 & 0.80 & 0.75 & 1.05 & [0.1, 5.0] & 1.64 & 0.99 \\
clip $[0.5,2.0]$               & 0.50 & 0.80 & 0.75 & 1.05 & [0.5, 2.0] & 1.64 & 0.50 \\
\bottomrule
\end{tabular}
\caption{Post-hoc sensitivity of weight distribution, computed on the first 30 AIME evaluation rollouts at $K{=}256$ from the Cue-GRPO evaluation (problems 1--30 of the 90-problem AIME set). The mean cluster count of 1.64 is lower than the training-run mean of 2.25 (Appendix~I), reflecting the lower structural diversity of correct AIME traces compared to the broader MATH training pool. $n_{\text{clust}}$: mean number of cue clusters per rollout group. $w_{\text{spread}}$: mean spread (max $-$ min) of positive-completion weights after floor.}
\label{tab:supp_sensitivity}
\end{table}

\noindent\textbf{Interpretation.} Increasing $\varepsilon$ makes the similarity graph sparser, raising the number of clusters and weight spread. $\alpha$ controls weight spread without changing cluster structure. $\tau$ shifts all positive weights additively and does not affect $w_{\text{spread}}$. Note that the $\tau{=}1.00$ row is a post-hoc floor-threshold sensitivity calculation and is distinct from the $\mu$-matched training ablation in Appendix~G, which sets $\delta{=}0$ and re-normalizes the post-filter positive weights to unit mean. The perturbations produce predictable changes without numerical instability; $\varepsilon$ and $\rho$ have the largest effect on cluster granularity.

\end{document}